\documentclass[reprint,amsmath,amssymb,aps,nofootinbib]{revtex4-1}

\usepackage{graphicx}
\usepackage{dcolumn}
\usepackage{bm}
\usepackage{hyperref}
\usepackage{gensymb}
\usepackage[super]{nth}
\usepackage{array}
\usepackage{xcolor}
\usepackage{float}
\usepackage{subcaption}
\usepackage{booktabs}
\usepackage[utf8]{inputenc}

\begin{document}

\preprint{APS/123-QED}

\title{PixSet : An Opportunity for 3D Computer Vision to Go Beyond Point Clouds \\ With a Full-Waveform LiDAR Dataset}

\author{Jean-Luc Déziel}
 \affiliation{LeddarTech}
\author{Pierre Merriaux}
 \affiliation{LeddarTech}
\author{Francis Tremblay}
 \affiliation{LeddarTech}
\author{Dave Lessard}
 \affiliation{LeddarTech}
\author{Dominique Plourde}
 \affiliation{LeddarTech}
 \author{Julien Stanguennec}
 \affiliation{LeddarTech}
 \author{Pierre Goulet}
 \affiliation{LeddarTech}
 \author{Pierre Olivier}
 \affiliation{LeddarTech}
 
\date{\today}

\begin{abstract}
Leddar PixSet is a new publicly available dataset (\url{dataset.leddartech.com}) for autonomous driving research and development. One key novelty of this dataset is the presence of full-waveform data from the Leddar Pixell sensor, a solid-state flash LiDAR. Full-waveform data has been shown to improve the performance of perception algorithms in airborne applications but is yet to be demonstrated for terrestrial applications such as autonomous driving. The \emph{PixSet} dataset contains approximately 29k frames from 97 sequences recorded in high-density urban areas, using a set of various sensors (cameras, LiDARs, radar, IMU, etc.) Each frame has been manually annotated with 3D bounding boxes.
\end{abstract}

\maketitle

\section{Introduction}\label{sec:introduction}

Autonomous vehicles (AVs) have the potential to transform how transportation is done for people and merchandise, while improving both safety and efficiency. In order to reach the highest levels of autonomy, one of the main challenges that AVs are currently facing is to leverage the data from multiple types of sensors, each of which has its own strengths and weaknesses. Sensor fusion techniques are widely used to improve the performance and robustness of computer vision algorithms. 

Nowadays, the best performing computer vision algorithms are neural networks that are optimized using a deep learning approach \cite{Lang_2019_CVPR, SVGA-Net, Shi_2020_CVPR}, which requires large amount of data. Multiple datasets have been made publicly available in order to boost research and development of such algorithms \cite{kitty,pitropov_canadian_2020,caesar2020nuscenes,sun2020scalability,houston2020one}. In particular, most of these datasets include data acquired with a LiDAR sensor (Light Detection and Ranging) which is considered to be an essential component of highly autonomous vehicles \cite{9000872,9173706}. 

In this paper, we present our contribution to this effort, the \emph{PixSet} dataset\footnote{Download link: \url{dataset.leddartech.com}}. What makes this new dataset unique is the use of a flash LiDAR and the inclusion of the full-waveform raw data, in addition to the usual point cloud data. The use of full-waveform data from a flash LiDAR has been shown to improve the performance of segmentation and object detection algorithms in airborne applications \cite{airborne, Shinohara_2020}, but is yet to be demonstrated for terrestrial applications such as autonomous driving. 

\begin{figure}
    \centering
    \includegraphics[trim=100 50 100 100, clip, width=1.0\columnwidth]{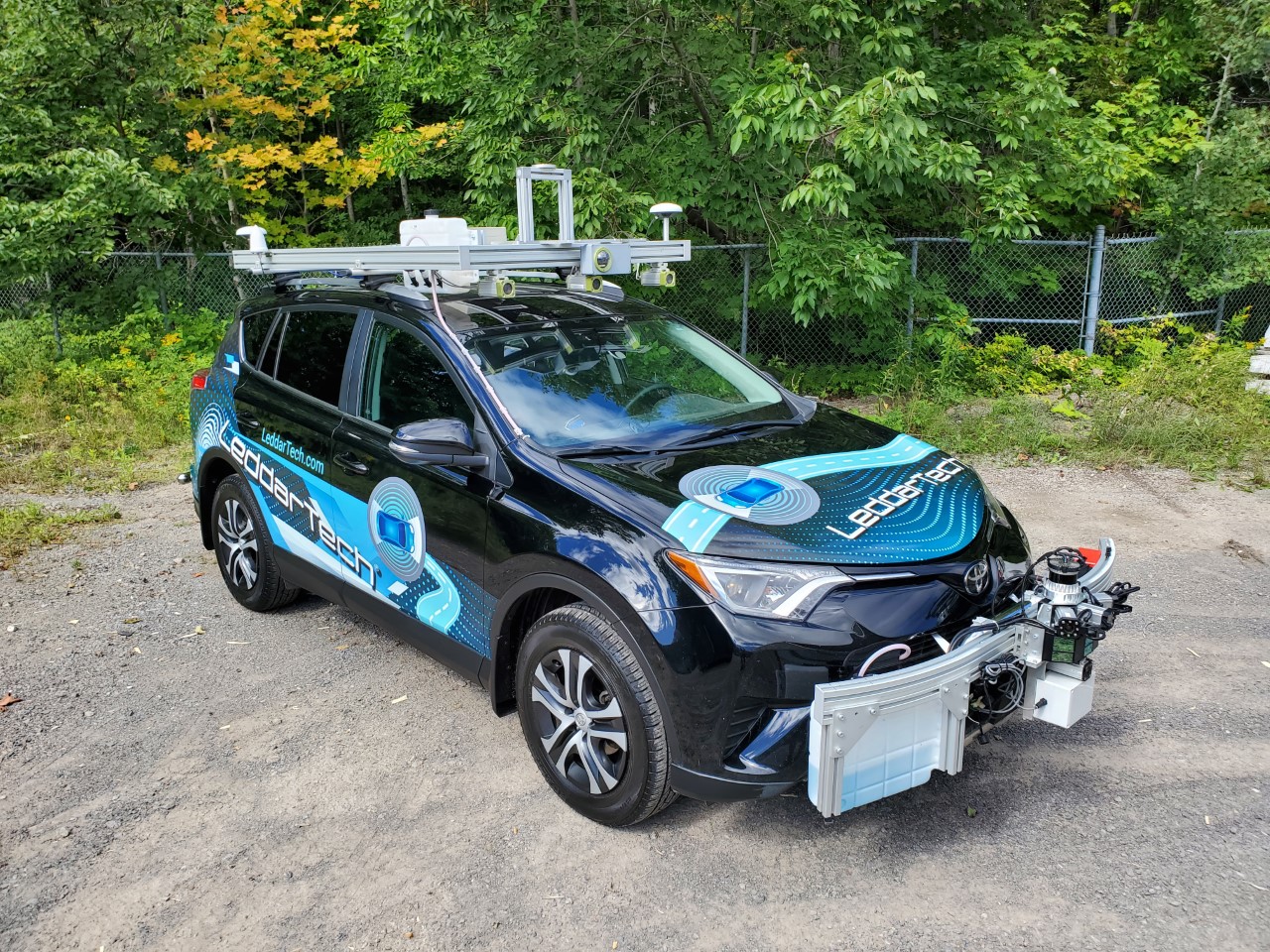}
    \caption{Instrumented vehicle (\emph{RAV4}) used for data acquisition.}
    \label{fig:rav4_figure}
\end{figure}

The \emph{PixSet} dataset contains 97 sequences, each averaging a few hundreds of frames, for a total of roughly 29000 frames. Each frame has been manually annotated with 3D bounding boxes. The sequences have been gathered in various environments and climatic conditions with the instrumented vehicle shown in Figure~\ref{fig:rav4_figure}. 

Our main contributions are summarized as follows: 
\begin{itemize}
  \item Introduce to the community a new dual LiDAR-type dataset using solid-state and mechanical LiDARs, with 3D bounding box annotation.
  \item Provide full-waveform data for the solid-state LiDAR.
  \item Trigger exteroceptive sensors to improve 3D bounding box annotation accuracy
  \item Provide API and dataset viewer to facilitate algorithm research and development.
\end{itemize}

The paper is organized as follows: sections \ref{sec:sensors} to \ref{sec:calibration} present the recording setup and conditions. Section \ref{sec:format} briefly describes how the data is stored and how to read/viewed it with an optional open source library we have developed for \emph{PixSet}. The waveform data is described in section \ref{sec:waveforms}. Section \ref{sec:annotations} describes the dataset annotations (3D bounding boxes). Then section \ref{sec:object_detection} provides baseline object detection results, and a brief conclusion is presented in section \ref{sec:conclusion}.

\section{Dataset}\label{sec:dataset}

The dataset is available at the link \url{dataset.leddartech.com}. The format of the data is discussed in section \ref{sec:format}.

\subsection{Sensors}\label{sec:sensors}

\begin{figure}
    \centering
    {\includegraphics[trim=10.8cm 3cm 11cm 4.2cm, clip, width=1.0\columnwidth]{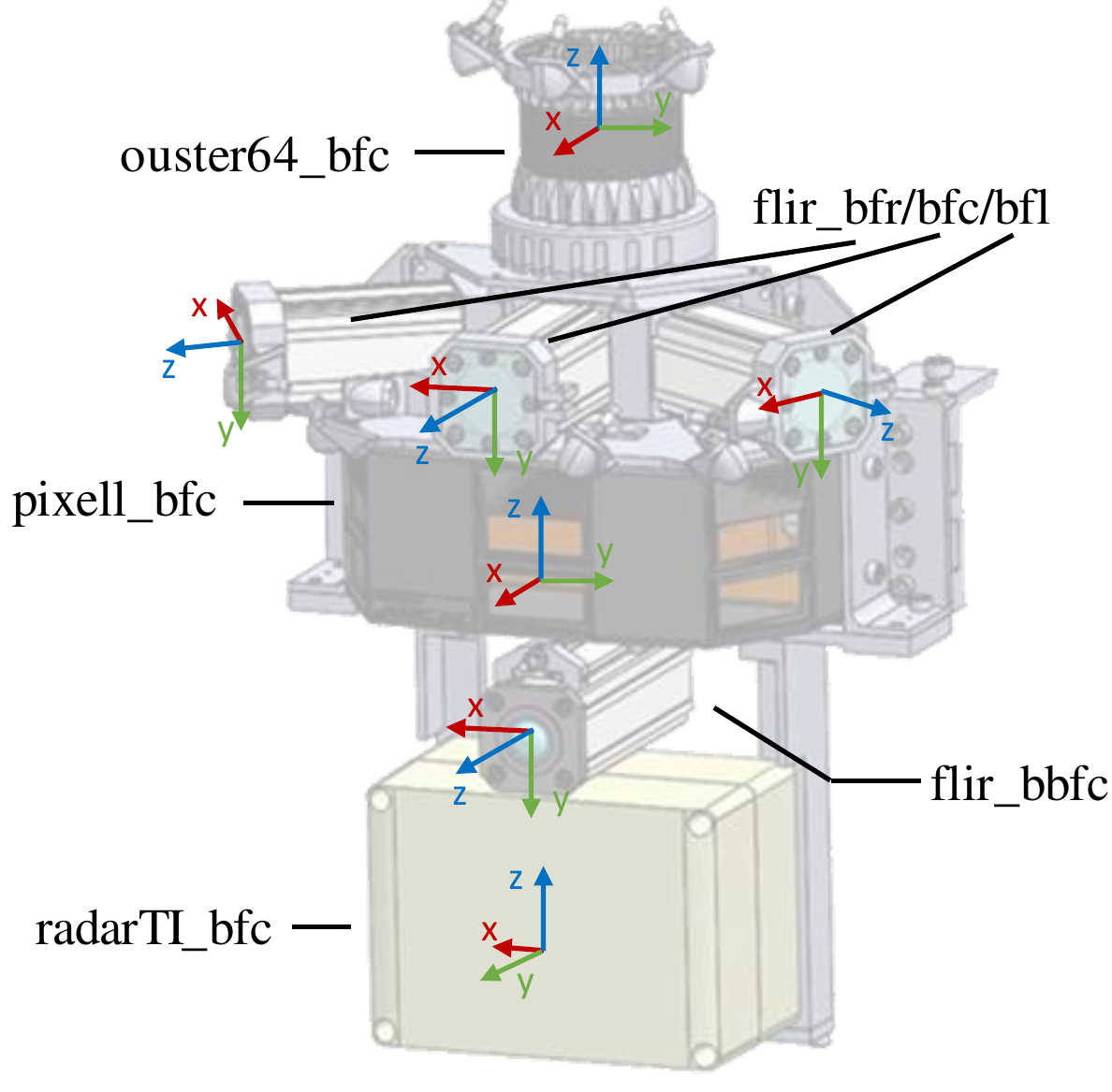}}
    \caption{Spatial configuration of the sensor stack on the front of the vehicle. Arrows show the coordinate system used by each sensor.  See Table \ref{table:1} for a description of the sensors.}
    \label{fig:pod_axis}
\end{figure}

The sensors used to collect the dataset were mounted on a car (see Figure \ref{fig:rav4_figure}) and are listed in Table \ref{table:1}. Most sensors (cameras, LiDARs and the radar) are positioned close to each other at the front of the car in a configuration shown in Figure \ref{fig:pod_axis}. This proximity is deliberate in order to minimize the parallax effect. The GPS antennas for the inertial measurement unit (IMU) are located on the top of the vehicle.

\begin{table}

\begin{tabular}{r m{5.5cm}   } 
\toprule \hline
\multicolumn{1}{c}{\emph{Sensor label}} & 
\multicolumn{1}{c}{\emph{Description}} 
\\ 
\midrule \hline
	pixell\_bfc & Leddar Pixell solid-state LiDAR with waveforms \\ 
	ouster64\_bfc & Ouster OS1-64 mechanical LiDAR \\ 
    flir\_bfl/bfc/bfr & 3 FLIR BFS-PGE-16S2C-CS cameras + 90\degree optics \\ 
	flir\_bbfc & FLIR BFS-PGE-16S2C-CS camera + Immervision panomorph 180\degree optic.\\ 
	radarTI\_bfc & TI AWR1843 mmWave radar \\ 
	sbgekinox\_bcc & SBG Ekinox IMU with RTK GPS dual antenna\\ 
	peakcan\_fcc & Toyota RAV4 CAN bus \\
\bottomrule \hline
\end{tabular}

\caption{List of sensors used for \emph{PixSet} dataset data acquisition.}
\label{table:1}
\end{table}

Most sensors are fairly standard and well known by the community, except for the Leddar Pixell. It is a solid-state (no moving parts) flash and full-waveform LiDAR (see section \ref{sec:waveforms} for a description of the waveforms). Its field of view is 180\degree horizontally and 16\degree vertically. While it has a relatively low resolution (96 horizontal channels by 8 vertical channels), it should provide more information per channel than a typical non-flash LiDAR sensor. Testing this hypothesis is one of the motivations to create this dataset.

\subsection{Time Synchronization}\label{sec:synchronization}

\begin{figure}
    \centering
    \includegraphics[trim=65 30 0 35, clip, width=1.0\columnwidth]{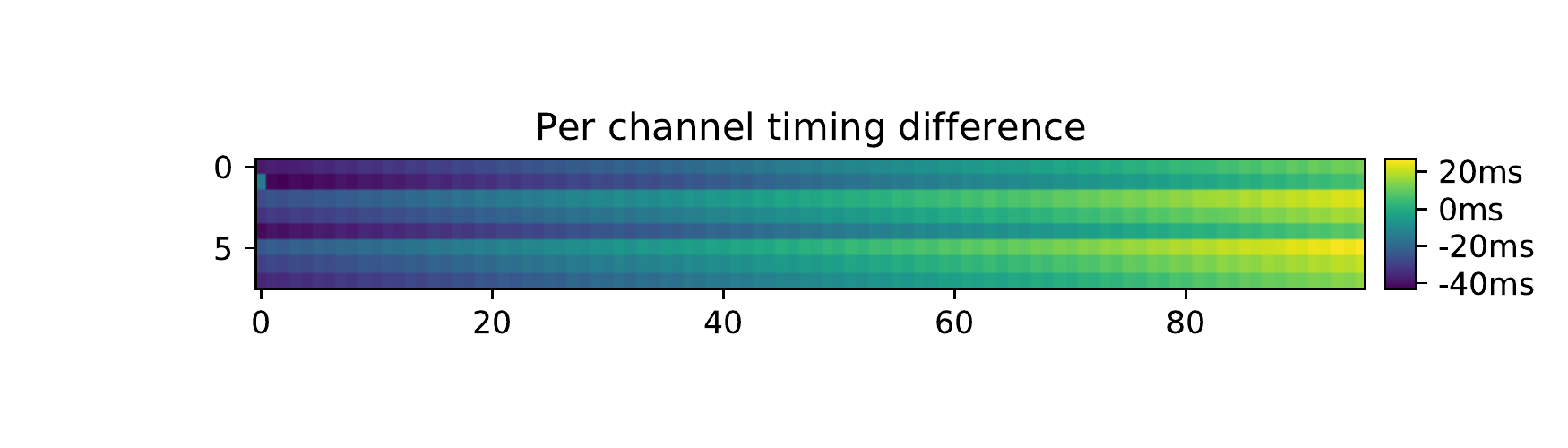}
    \caption{Average difference between the timestamp of each channel of the Pixell sensor and the timestamps of the projected overlapping OS1-64 detections.}
    \label{fig:ts_diff_ouster_pixell}
\end{figure}

One of the objectives of \emph{PixSet} was to obtain the highest accuracy possible in the positioning of the bounding boxes for data annotation.  Since the environment is dynamic, combining the data from multiple sensors will yield the best result when the sensors are gathering data simultaneously.  This was achieved by triggering each frame acquisition from a single periodic signal source generated by the OS1-64 mechanical LiDAR. The Pixell LiDAR and the 4 cameras are triggered as such, at a frequency of 10 Hz. The other sensors (radar, IMU and the vehicle CAN bus) are not triggered, but are recording at much higher frame rates which minimizes the timing errors. 
The sensor triggering is used to minimize the acquisition time difference between extrinsic sensors. In Figure \ref{fig:ts_diff_ouster_pixell}, the average offset differences were measured for every channel of the Pixell sensor, which is 40 ms at most.

A timestamp is associated with each frame of each sensor, or each point for LiDAR. To make sure that all sensors are precisely synchronized, we run a server that collects the time from the GPS antenna (PTP protocol) and connect all sensors to that server to synchronize their internal clocks. The only exceptions are the radar and the vehicle CAN bus for which the timestamps are assigned at data reception in the computer (the computer's clock is also synchronized with the same time server). Moreover, the quality of the synchronization is always monitored in real time by comparing the trigger signal to a separate time measurement from the IMU. The typical measured synchronization error was 6 $\mu$s for cameras and 350 $\mu$s for Pixell.

One more thing that was considered is the fact that LiDARs are not global shutter sensors, in contrast with cameras. What is referred to as a "frame" for a LiDAR, or a single complete scan of the field of view, is not measured all at once. For example, a typical mechanical LiDAR such as the OS1-64 is simultaneously measuring a single vertical line, which is continuously rotating during 100ms. Thanks to accurate timestamping LiDAR and IMU, the motion of the car during the frame acquisition time can be compensated. The API (section~\ref{sec:format}) provides this functionality to algorithm developers.

\subsection{Calibration}\label{sec:calibration}

\begin{figure}
    \centering
    \includegraphics[trim=2 0 0 0, clip, width=1.0\columnwidth]{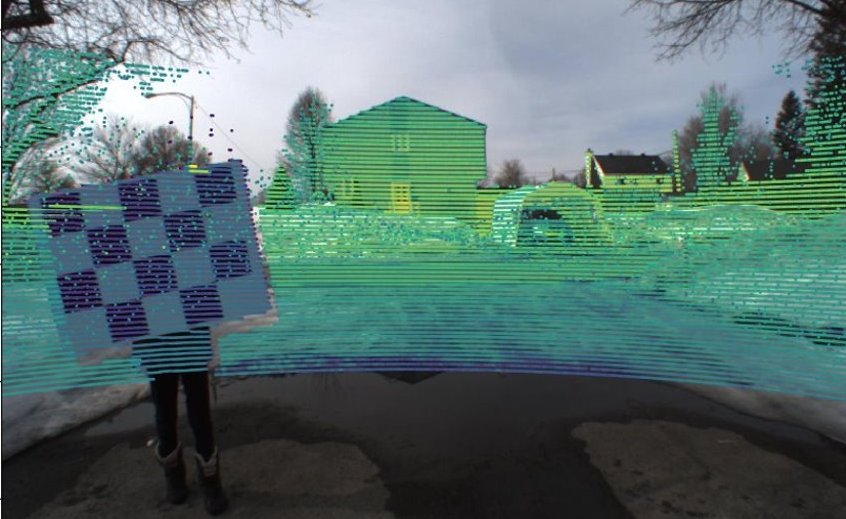}
    \caption{Projection of an OS1-64 point cloud in a camera image, using the calibration matrices provided with the dataset.}
    \label{fig:ouster_camera_projection}
\end{figure}

Sensor projection is mandatory for many algorithms like sensor fusion. To obtain an accurate projection (Figure \ref{fig:ouster_camera_projection}), a calibration must be performed, which can be done by multiple methods. The following describes the methods chosen to obtain the calibration matrices included in the dataset. 

To compensate for the camera lens distortion, a set of images of a chessboard was gathered for each camera and the openCV library that calculates the desired matrices from these images was used. For the Pixell sensor, the direction of each channel must be known in order to position each detection in 3D space. Thankfully, these directions are calibrated at the factory with specialized equipment and provided in the dataset.

Next, the extrinsic calibration matrices are also needed to change the referential of the coordinate system from/to any pair of sensors. The extrinsic calibration between pairs of cameras is obtained, again, by using the openCV library to extract the 3D coordinates of the corners of a chessboard in multiple images. A closed loop optimization method was closed to minimize the transformation matrices. A similar method is used for the extrinsic calibration between the OS1-64 LiDAR and the cameras, by first extracting the 3D coordinates of the corners of a chessboard in each sensor and then solving the transformation matrices with the perspective-n-point method. The extrinsic calibration between both LiDARs (Pixell and OS1-64) is obtained by averaging multiple matrices, each obtained by using the iterative closest point (ICP) method with pairs of simultaneous scans from both sensors. The calibration between the radar and the OS1-64 LiDAR is obtained similarly, except that only the high-intensity detections from a specific metallic target are accounted for. Finally, the calibration between the OS1-64 LiDAR and the IMU is obtained by minimizing plane thickness of multiple sequential point clouds in the world coordinates provided by the IMU. The pairs of sensors that have not been mentioned are obtained by combining other matrices (i.e. the Pixell to IMU transformation matrix is obtained by multiplying the matrices for Pixell to OS1-64 and for OS1-64 to IMU).

\subsection{Recording Conditions}\label{sec:recording}

\begin{figure*}
  \centering
  \includegraphics[trim=0 0 0 0, clip, width=1.0\textwidth]{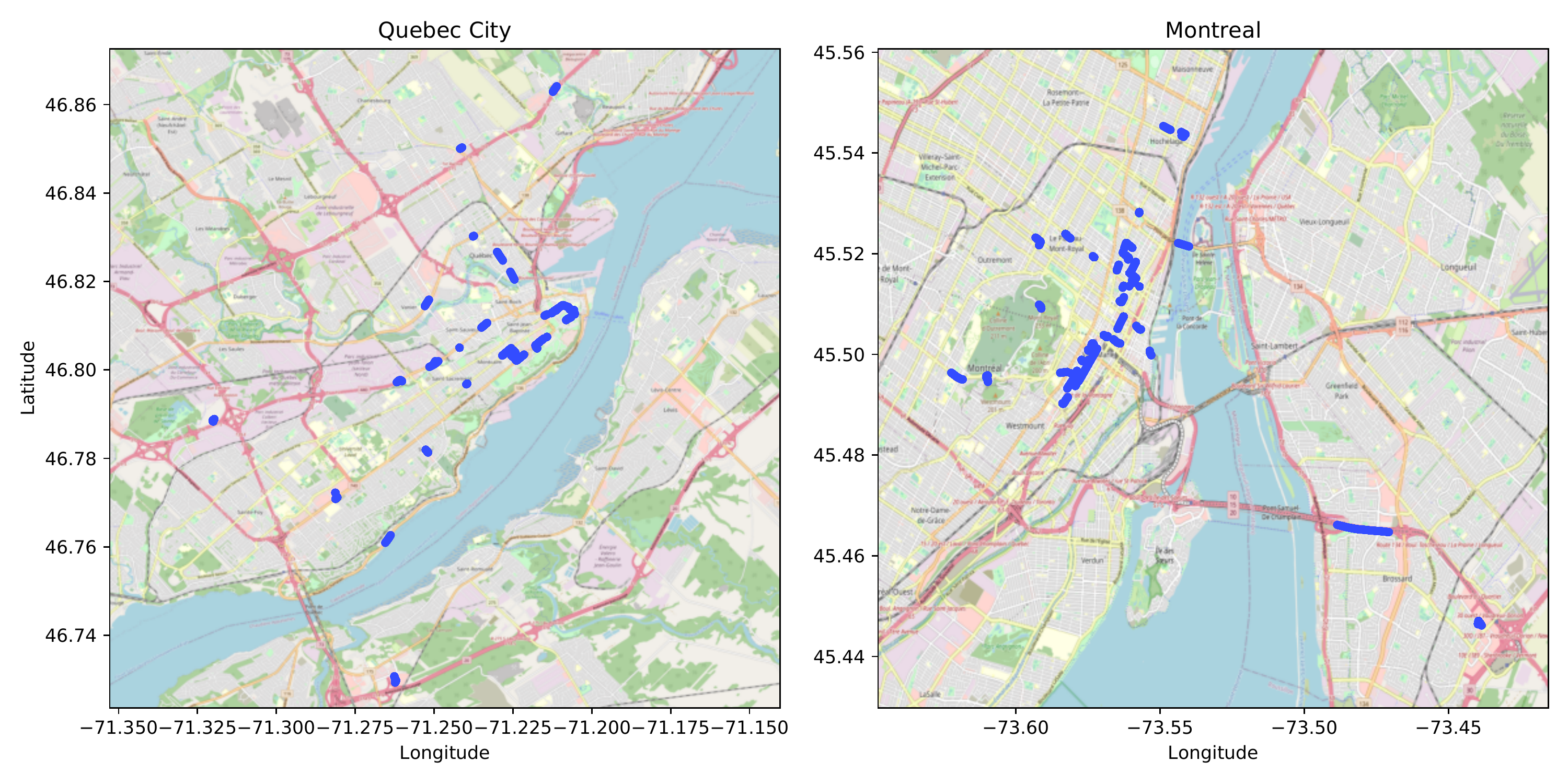}
  \caption{Locations of the recorded data.}
  \label{fig:map_data}
  \end{figure*}

\begin{figure*}
\centering
\includegraphics[trim=0 10 0 0, clip, width=1.0\textwidth]{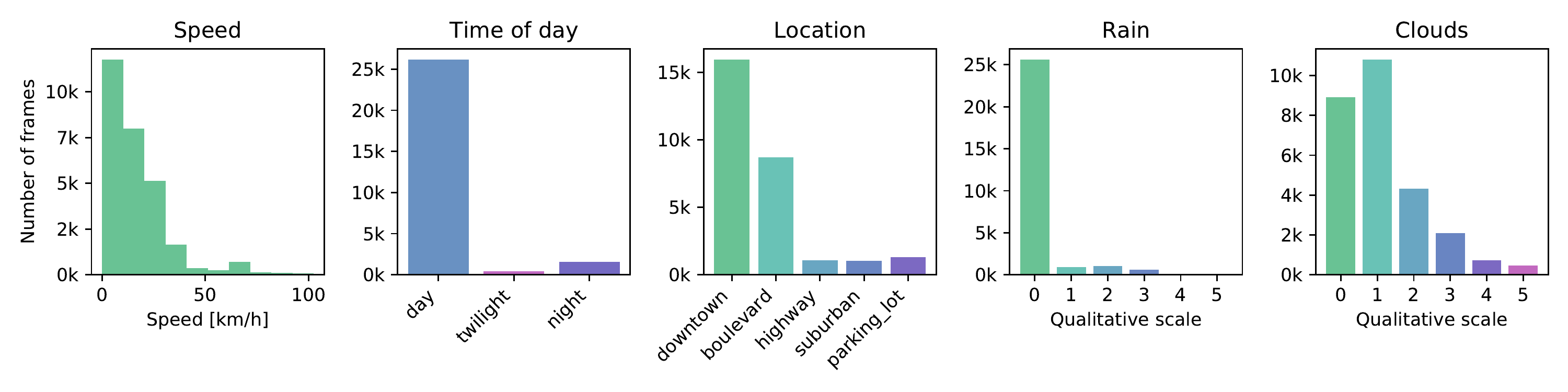}
\caption{Summary of recording conditions.}
\label{fig:driving_conditions}
\end{figure*}

\begin{figure*}
\centering
\includegraphics[width=0.9\textwidth]{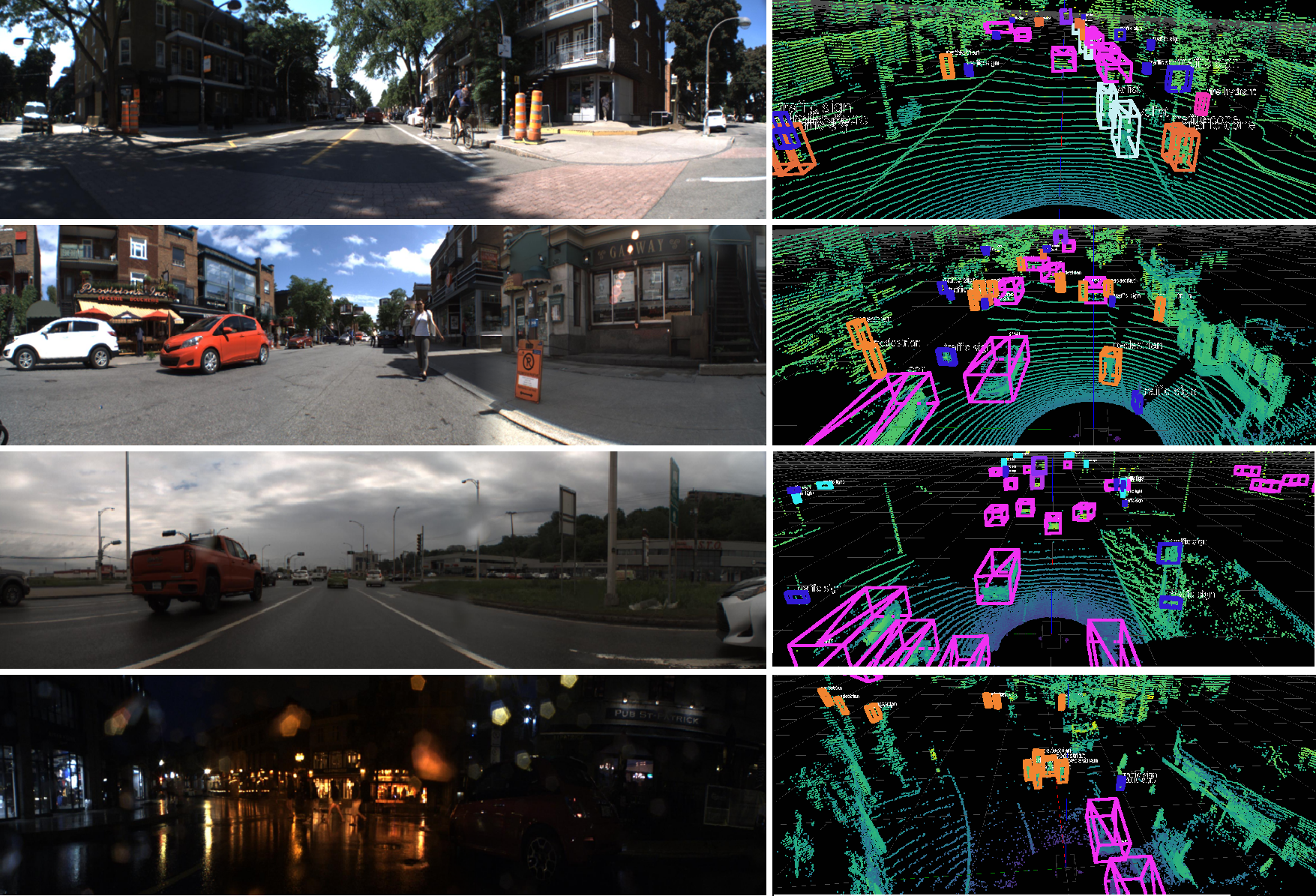}
\caption{\emph{PixSet} dataset overview of the variety of scenes and environmental conditions with samples from the cameras and the OS1-64 LiDAR with 3D boxes annotation.}
\label{fig:Pixset}
\end{figure*}

The \emph{PixSet} dataset has been recorded in Canada, in both Quebec City and Montreal, during the summer of 2020. Locations are shown in Figure \ref{fig:map_data}. A summary of the recording conditions is presented in Figure \ref{fig:driving_conditions}. Most of the dataset has been recorded in high-density urban environments, such as downtown areas where pedestrians are almost always present and on boulevards where the car density is high. Most of the dataset was recorded during day time with dry weather, but a few thousand frames were recorded at night and/or while raining, providing a great variety of situations with real-world data for autonomous driving. Figure \ref{fig:Pixset} showcases a few samples taken from the dataset.

\subsection{Dataset Format and Access}\label{sec:format}

Each of the 97 sequences of the dataset is contained in an independent directory. Each of them contains the following elements: (i) a configuration file named \textit{platform.yml} that contains all the required parameters to read and process the data from the optional library provided along with the dataset (see below for details). (ii) An \textit{intrinsics} directory that contains the calibration matrices for each camera (lens distortion compensation, see Section \ref{sec:calibration}). (iii) An \textit{extrinsics} directory that contains the extrinsic calibrations (4x4 affine transformation matrices) for the pairs of sensors that have been calibrated. (iv) A \emph{zip} file for each source of data (i.e. images from a camera).

A given \emph{zip} file is named after the label of the sensor (see Table \ref{table:1}, with \textit{pixell\_bfc} for the Pixell for example) and the type of data, which can be multiple for a single sensor. The raw waveforms from Pixell are stored in the \textit{pixell\_bfc\_ftrr.zip} file and the detections that have been processed from these waveforms are stored in the \textit{pixell\_bfc\_ech.zip} file. In each of these files, a series of \emph{pickle} files (or a \emph{.jpg} for camera images) are stored, each of them containing the data of a single frame (a full LiDAR scan or a single camera image). Along the raw data, a \textit{timestamps.csv} file is also stored,  containing all timestamps.

Along with the dataset, we also provide an API to a Python library that was developed to read, process and view the data\footnote{\url{https://github.com/leddartech/pioneer.das.api} \\ \url{https://github.com/leddartech/pioneer.das.view}}. This tool can perform several of the most complex and common operations such as referential transformations, time synchronization of the different sources of data, motion compensation for point clouds and much more.

\subsection{Waveforms}\label{sec:waveforms}

\begin{figure}
\centering
\includegraphics[trim=0 0 0 0, clip, width=1.0\linewidth]{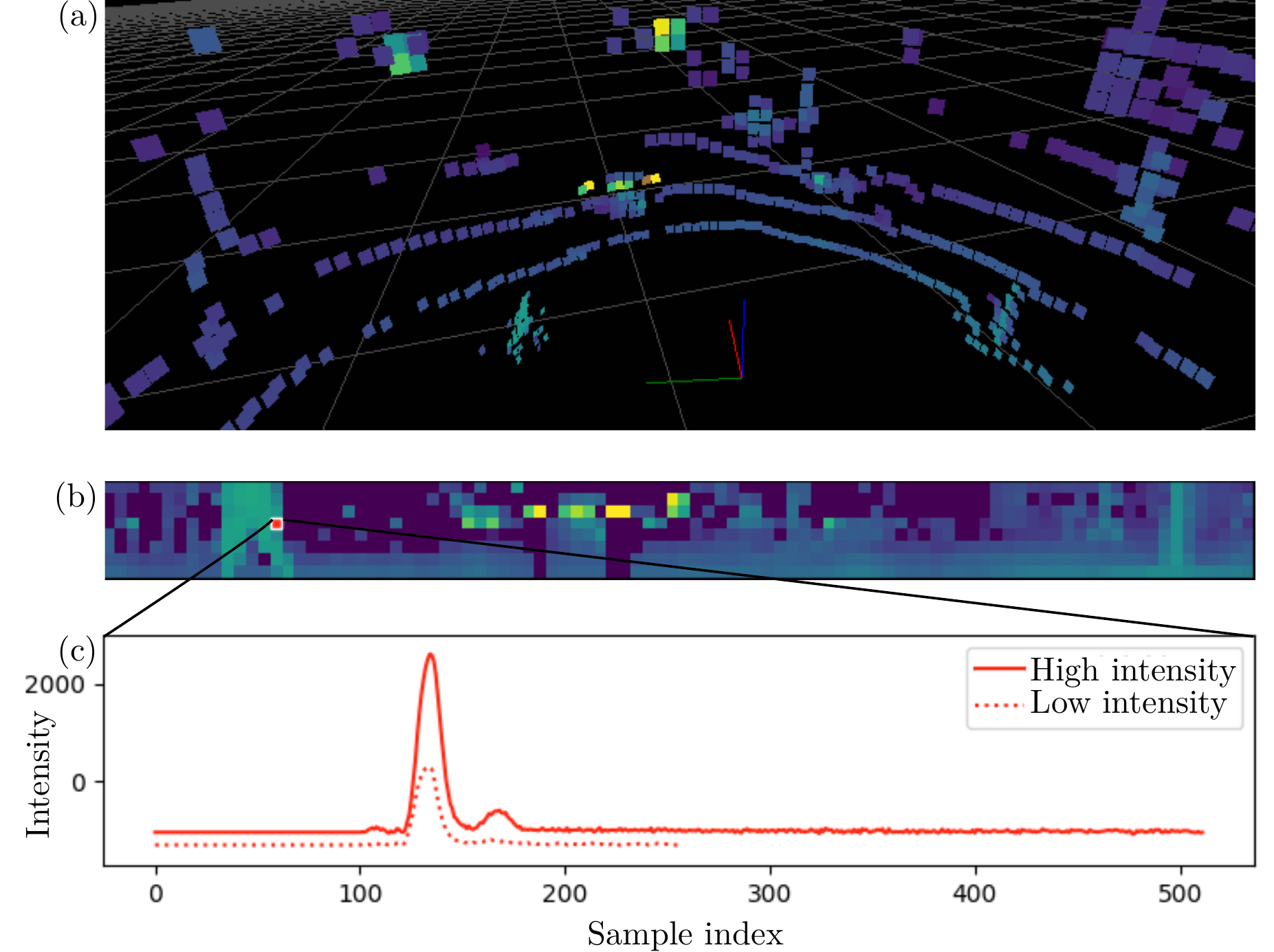}
\caption{(a) 3D view of detections provided by the Pixell sensor. (b) 2D front view of the amplitudes from the same detections as in (a). (c) Example of raw waveforms obtained from a single Pixell channel. (See text for details).}
\label{fig:waveforms}
\end{figure}

Typically, a LiDAR provides data in the form of point clouds, a collection of three-dimensional coordinates. Usually, an intensity value is also attached to each point. Point clouds are not the raw data measured by the sensor but are rather processed from the waveforms. A waveform is the measured intensity as a function of time after the emission of the laser pulse from the LiDAR (See Figure \ref{fig:waveforms} for an example). A point is the result of a peak found in a waveform. While the information about the positions and the amplitudes of the peaks is preserved in the point cloud, all additional information about the shape of the waveform is lost.

The Pixell sensor measures a set of two waveforms for each channel of each frame. The first waveform is gathered after the emission of a high-intensity laser pulse and the second one after the emission of a laser pulse with a quarter of the intensity. This effectively increases the dynamic range of the sensor, because the high-intensity waveforms have a tendency to saturate the sensor when looking at close and/or reflective targets. This is analogous to high-dynamic range imagery with cameras that can be achieved by combining images with different exposure levels.

The waveforms are sampled at an 800 MHz rate. The high-intensity waveforms contain 512 samples, while the low-intensity ones contain 256 samples. One more important aspect to keep in mind is that waveforms from different channels have slightly different time offsets with respect to each other. These offsets are calibrated and compensated for in the provided point clouds, but not for the raw waveforms. The offsets are provided with the dataset and the waveforms can easily be adjusted (by linear interpolation). The API provided (see end of section \ref{sec:format}) contains multiple functions to deal with waveform processing, including time realignment.

\section{Annotations}\label{sec:annotations}

Each frame ($\sim$29k) of the dataset has been manually annotated with 3D bounding boxes (Figure~\ref{fig:Pixset}). The annotators have been provided with the point clouds from both the Pixell and OS1-64 LiDARs, merged into the same referential (see section \ref{sec:calibration}). Motion compensation have been applied to both point clouds (see end of section \ref{sec:format}). Images from the cameras \emph{flir\_bfl, flir\_bfc} and \emph{flir\_bfr} (see Figure \ref{fig:pod_axis}) along with their calibration data have also been provided to be able to project the point clouds into the images to help identity objects.

\begin{figure*}
  \centering
  \begin{minipage}{.45\linewidth}
    \subcaptionbox{Total number of annotated objects per category. Interestingly, the relative amounts seem to follow Zipf's law \cite{zipf}.}
      {\includegraphics[trim=0 10 0 0, clip, width=\linewidth]{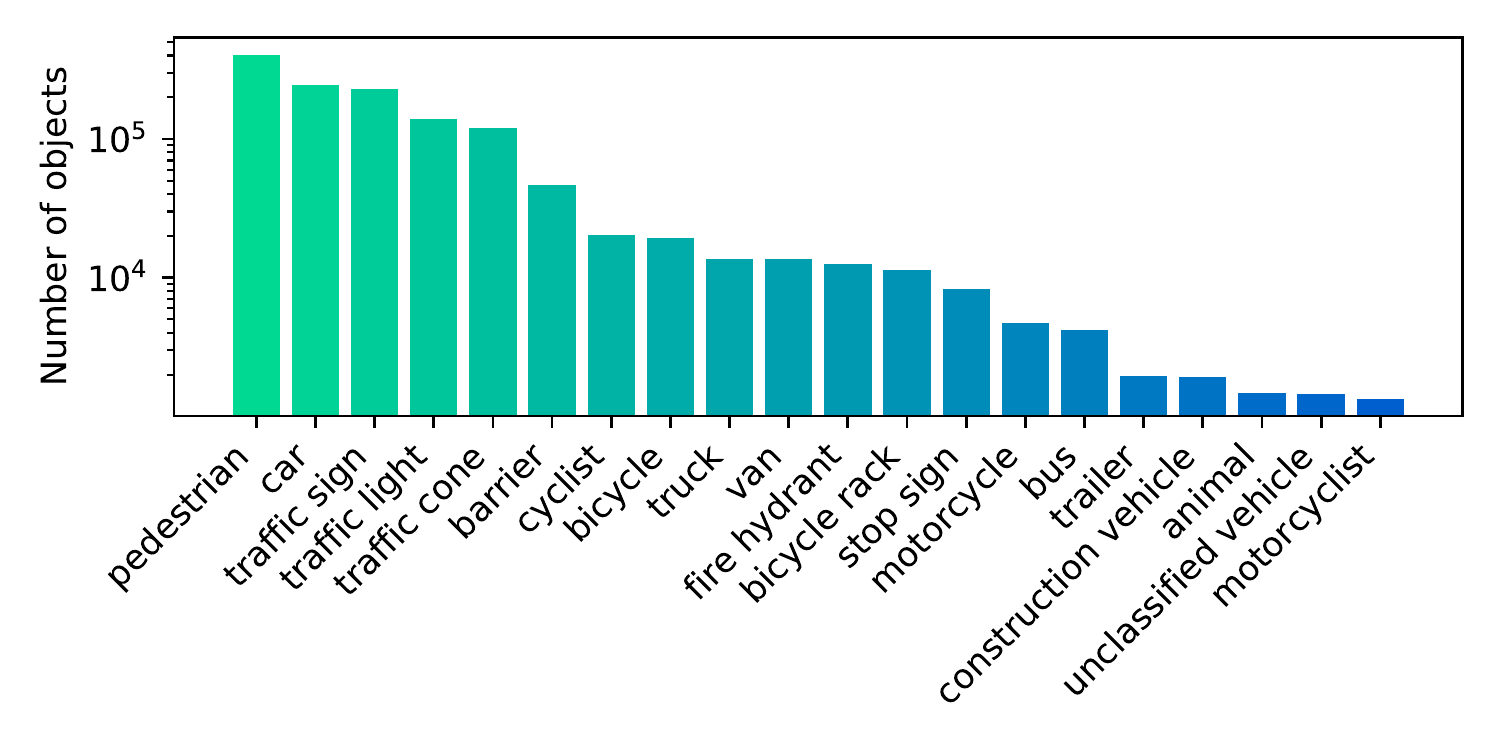}}
    \subcaptionbox{Distribution of the distances and orientations of cars, pedestrians and cyclists.}
     {\includegraphics[trim=10 0 10 -40, clip, width=\linewidth]{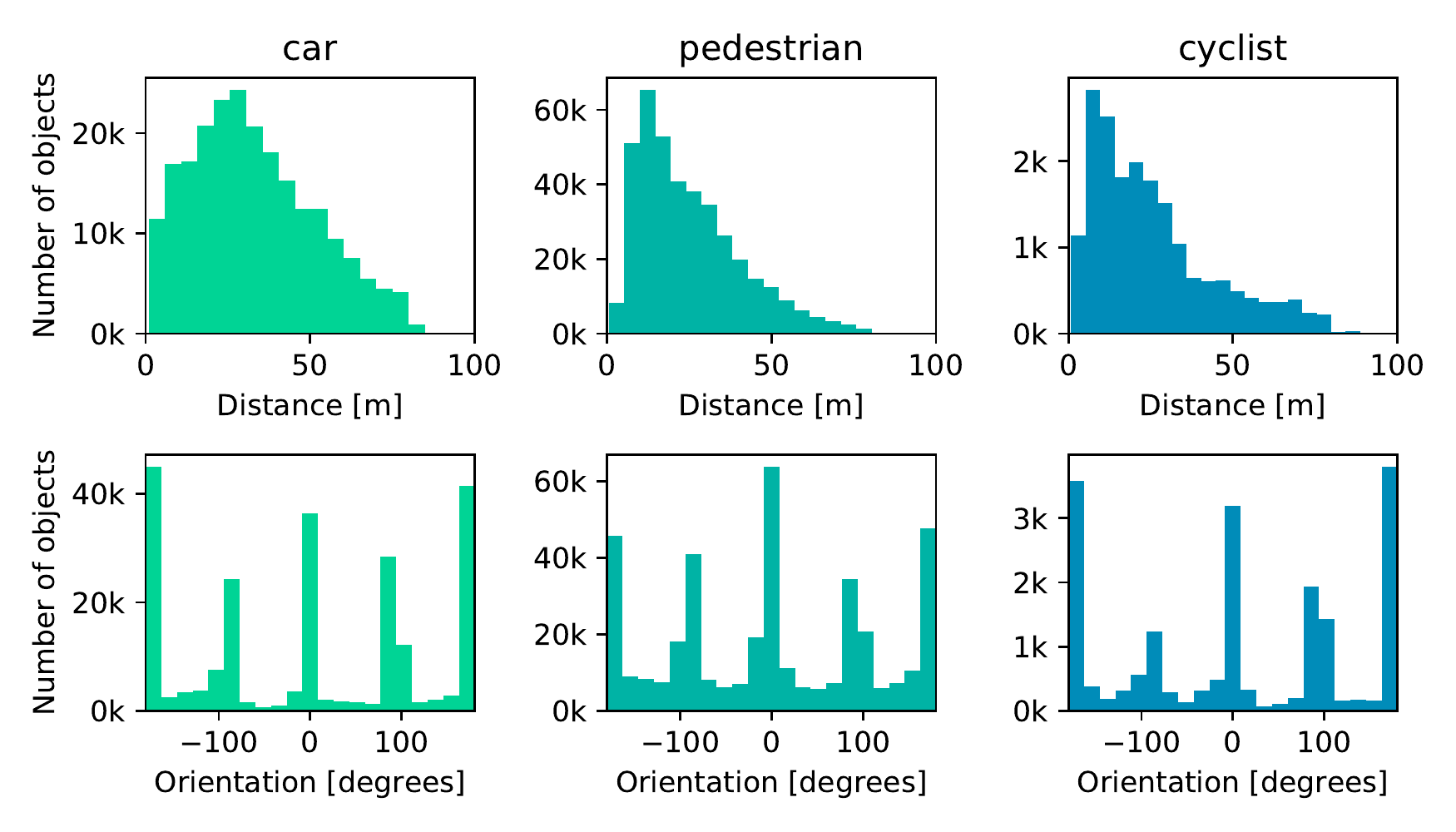}}%
  \end{minipage}%
  \hfill
  \begin{minipage}{.45\linewidth}
    \subcaptionbox{Relative amount of each attribute values found in the dataset for cars, pedestrians and cyclists. Occlusion refers to a fraction of the object hidden behind other things. Truncation refers to a fraction of the object that is outside the LiDARs' field of view. Take note that an occlusion/truncation value of 0 means $0\%$, a value of 1 means $<50\%$ and a value of 2 means $>50\%$.}
      {\includegraphics[trim=0 20 0 0, clip, width=\linewidth]{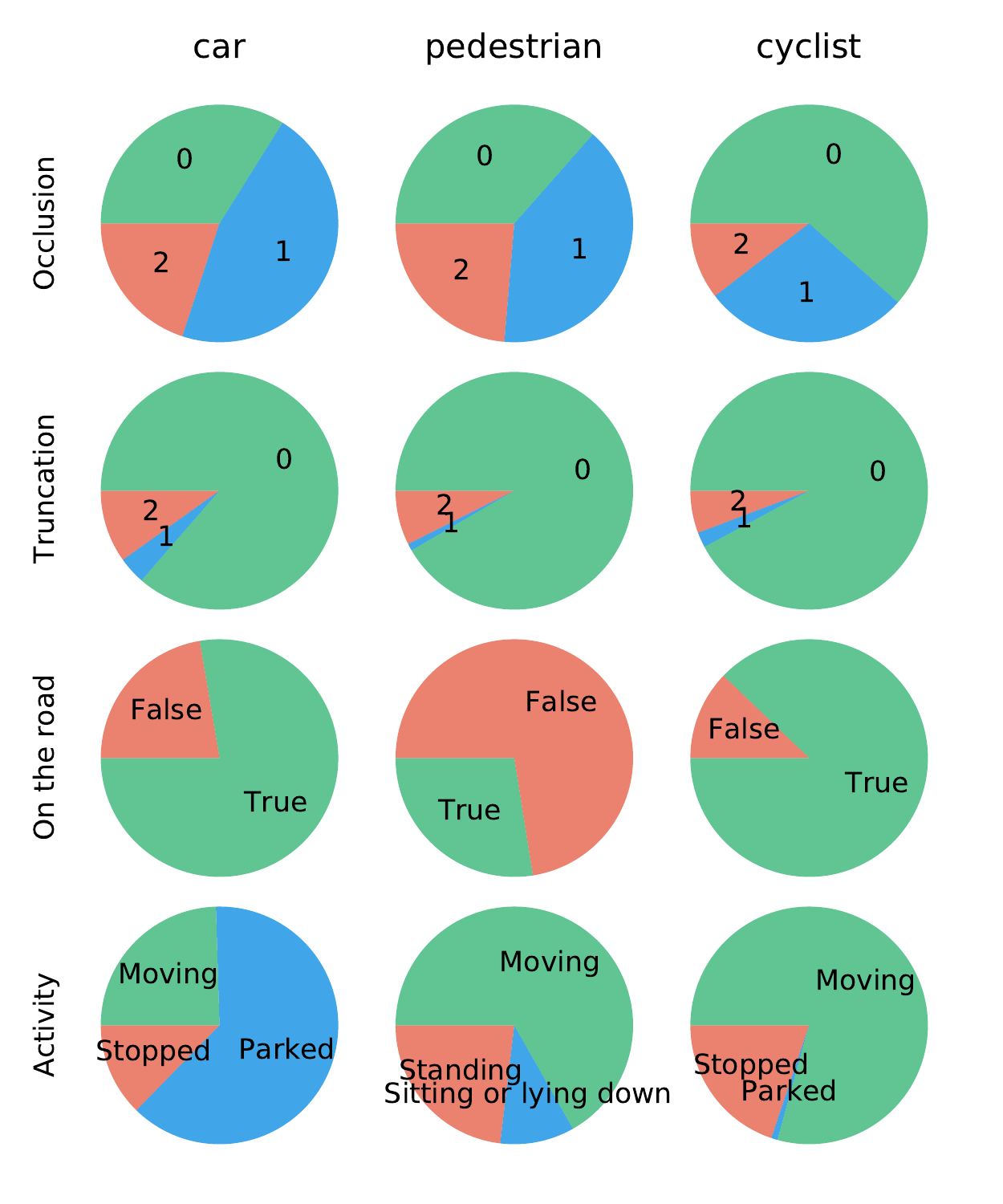}}%
  \end{minipage}%
  \caption{Overview of data annotations.}\label{fig:annotations}
\end{figure*}

Each bounding box is associated with one physical object and describes its central position (in the Pixell sensor's referential), its dimensions and orientations with respect to all three axes, its category (car, pedestrian, etc.), an inter-frame persistent identification number and a set of attributes such as an occlusion level. The complete list of the 20 categories and the number of annotated objects per category are shown in Figure \ref{fig:annotations}(a). The distributions or distances and orientations of the bounding boxes for a few categories are shown in Figure \ref{fig:annotations}(b). An overview of the attributes for some categories is also presented in Figure \ref{fig:annotations}(c).

A notable feature of the \emph{PixSet} dataset is the variable size of the bounding boxes for pedestrians. It is common to use a fixed size for a given object instance for the whole duration of a sequence. It makes sense for cars, or any solid object that has a fixed size and shape, but leads to complications for pedestrians that are deformable objects. For example, if a pedestrian extends their arm for a few seconds, using a fixed bounding box size is problematic. Should the arm be ignored or included in the box, and an oversized bounding box maintained for the remainder of the sequence? Thus, bounding boxes for pedestrians in the \emph{PixSet} dataset have variable sizes, adjusted frame per frame, to avoid these issues.

\section{Object Detection}\label{sec:object_detection}

This section presents results from an object detection algorithm in order to provide a baseline for the perception performance of the Pixell sensor. It describes the model and the metrics at a coarse level, while the source code and all details can be found in \cite{objectdetectionpixell}. 

\paragraph{Preprocessing.}
The raw data is prepared and preprocessed by using the library that was made publicly available (See section \ref{sec:format}). The steps performed by this tool are the following: (i) match the synchronized data from Pixell, annotations and the IMU for egomotion. (ii) Convert raw detection data of the Pixell in point clouds. (iii) Apply motion compensation to the point cloud (See end of section \ref{sec:synchronization}). 

Then, data augmentation is performed using the following methods: (i) insertion of bounding boxes and contained points from other frames (up to 5 cars, 5 pedestrians and 5 cyclists); (ii) With a 50\% probability, flip the left/right axis. (iii) Random translations along all three axes within the values $x \in [-3,0]$, $y \in [-3,3]$, $x \in [-0.5,0.5]$ (in meters, see Figure \ref{fig:pod_axis} for the coordinate system). (iv) Random rotation (\emph{yaw}~$\in[-5\degree,5\degree]$). (v) Random scaling of the scene by a factor $f\in[0.95,1.05]$. The translations, rotation and scaling are applied to the entire frame, using the position of the sensor as the origin.

\paragraph{Neural network.}
The neural network is composed of three parts: the encoder, the backbone and the detection head. The encoder is similar to the one from PointPillars \cite{Lang_2019_CVPR}, except that the single encoding layer is replaced by a dense block inspired by DenseNet \cite{8099726} (4 layers, with a growth parameter of 16). This results in a 2D bird's eye view of a set of encoded features. The features then go through the backbone of the network, also a dense block, with 60 layers and a growth parameter of 16. The detection head is a final dense block of 5 layers and a growth parameter of 7. The detection head is inspired by CenterNet \cite{centernet} and produces bird's eye view heat maps where local maxima are interpreted as objects. Non-maximum suppression is not necessary with this method so it is not used. 

The neural network has a total of 5.2M parameters. The model was trained for 100 epochs with the Adam optimizer \cite{DBLP:journals/corr/KingmaB14}. The learning rate was exponentially decaying from 0.001 down to 0.000174 between each epoch. A test set ($\sim$3k frames) was removed from the training data. The sequences that were kept for the test set are parts 1, 9, 26, 32, 38 and 39 (these part numbers are in the name of each downloadable directory of the dataset). 

\begin{figure*}
\centering
\includegraphics[trim=0 10 0 0, clip, width=1.0\textwidth]{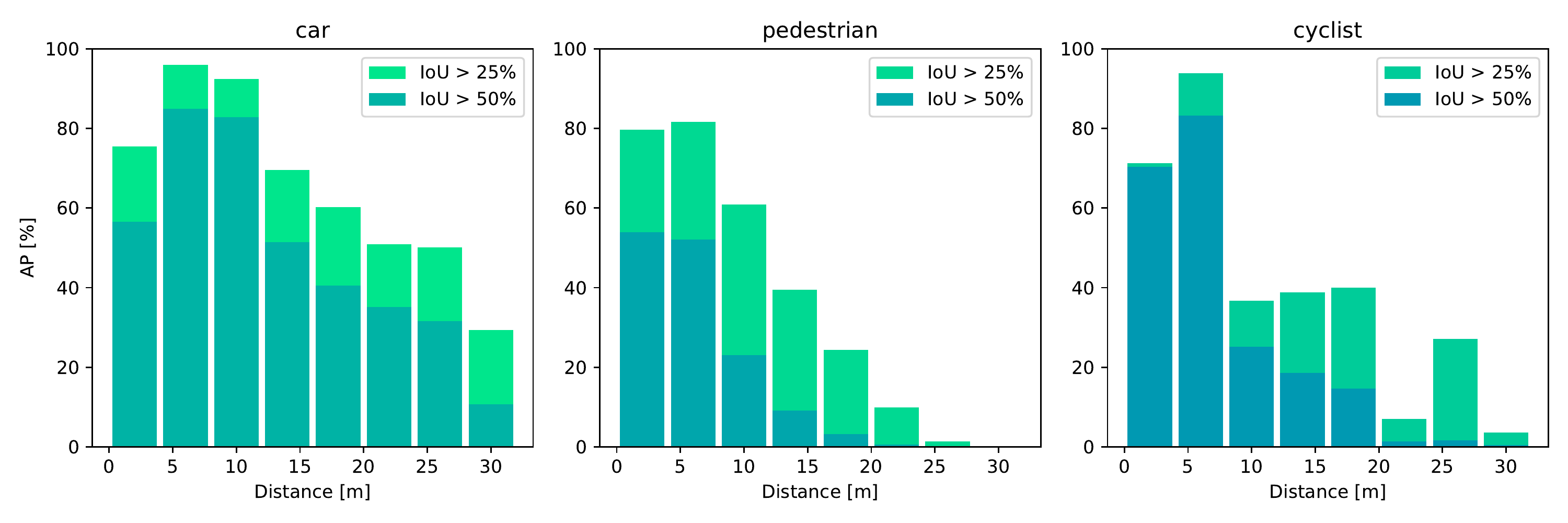}
\caption{Average precision (AP) as a function of distance for cars, pedestrians and cyclists found in the test set (parts 1, 9, 26, 32, 38 and 39 of the dataset). To be considered a true positive, a prediction must have an IoU (intersection over union) larger than the thresholds indicated in the legends.}
\label{fig:object_detection}
\end{figure*}

\paragraph{Metrics.}
The most popular metric to evaluate the performance of an object detection model is the average precision (AP). This metric is calculated as a function of distance, as shown in Figure \ref{fig:object_detection} (See table \ref{tab:ap} for the global AP values). This better shows how a sensor can perform quite well at short range and at what distance the model's predictions can be trusted. Admittedly, the Pixell detection capabilities seem to rapidly drop beyond 10-15 meters, but it is not meant to be used as the only sensor on an AV. These results are mostly presented as a reference baseline. Future work will focus on the performance of a fused sensor stack and the potential improvements of including full-waveform data from the Pixell.

\begin{table}
    \centering
    \begin{tabular}{c|c|c|c}
        IoU threshold & Car AP & Pedestrian AP & Cyclist AP \\ \hline
        25\% & 68.33\% & 37.15\% & 35.87\% \\
        50\% & 52.04\% & 14.13\% & 19.69\%
    \end{tabular}
    \caption{Average precision (AP) results. The evaluation only account for bounding boxes found within 32 meters from the Pixell sensor.}
    \label{tab:ap}
\end{table}

A few notes on the metric calculation: a predicted bounding box is considered a true positive if its three-dimensional intersection over union (IoU) is above a certain threshold (indicated in the figures) and if the classification of the bounding box is correct. Moreover, there can be only one true positive prediction per ground truth bounding box (duplicates are false positives).

\section{Conclusion}\label{sec:conclusion}

This work presented our contribution, the \emph{PixSet} dataset, to the collective effort of developing safe autonomous vehicles. We believe that there is a great potential to improve further the perception algorithms by leveraging the raw data from the full-waveforms provided by the Pixell flash LiDAR. We have also provided a baseline for 3D object detection performance from the Pixell point clouds.

\bibliography{bibliography}

\end{document}